# RL$^2$: Fast Reinforcement Learning via Slow Reinforcement Learning

**Yan Duan**$^{\dagger\ddagger}$, **John Schulman**$^{\dagger\ddagger}$, **Xi Chen**$^{\dagger\ddagger}$, **Peter L. Bartlett**$^{\dagger}$, **Ilya Sutskever**$^{\ddagger}$, **Pieter Abbeel**$^{\dagger\ddagger}$
$^{\dagger}$ UC Berkeley, Department of Electrical Engineering and Computer Science
$^{\ddagger}$ OpenAI
{rocky,joschu,peter}@openai.com, peter@berkeley.edu, {ilyasu,pieter}@openai.com

## Abstract

Deep reinforcement learning (deep RL) has been successful in learning sophisticated behaviors automatically; however, the learning process requires a huge number of trials. In contrast, animals can learn new tasks in just a few trials, benefiting from their prior knowledge about the world. This paper seeks to bridge this gap. Rather than designing a "fast" reinforcement learning algorithm, we propose to represent it as a recurrent neural network (RNN) and learn it from data. In our proposed method, RL$^2$, the algorithm is encoded in the weights of the RNN, which are learned slowly through a general-purpose ("slow") RL algorithm. The RNN receives all information a typical RL algorithm would receive, including observations, actions, rewards, and termination flags; and it retains its state across episodes in a given Markov Decision Process (MDP). The activations of the RNN store the state of the "fast" RL algorithm on the current (previously unseen) MDP. We evaluate RL$^2$ experimentally on both small-scale and large-scale problems. On the small-scale side, we train it to solve randomly generated multi-armed bandit problems and finite MDPs. After RL$^2$ is trained, its performance on new MDPs is close to human-designed algorithms with optimality guarantees. On the large-scale side, we test RL$^2$ on a vision-based navigation task and show that it scales up to high-dimensional problems.

## 1 Introduction

In recent years, deep reinforcement learning has achieved many impressive results, including playing Atari games from raw pixels (Guo et al., 2014; Mnih et al., 2015; Schulman et al., 2015), and acquiring advanced manipulation and locomotion skills (Levine et al., 2016; Lillicrap et al., 2015; Watter et al., 2015; Heess et al., 2015; Schulman et al., 2015; 2016). However, many of the successes come at the expense of high sample complexity. For example, the state-of-the-art Atari results require tens of thousands of episodes of experience (Mnih et al., 2015) per game. To master a game, one would need to spend nearly 40 days playing it with no rest. In contrast, humans and animals are capable of learning a new task in a very small number of trials. Continuing the previous example, the human player in Mnih et al. (2015) only needed 2 hours of experience before mastering a game. We argue that the reason for this sharp contrast is largely due to the lack of a good prior, which results in these deep RL agents needing to rebuild their knowledge about the world from scratch.

Although Bayesian reinforcement learning provides a solid framework for incorporating prior knowledge into the learning process (Strens, 2000; Ghavamzadeh et al., 2015; Kolter & Ng, 2009), exact computation of the Bayesian update is intractable in all but the simplest cases. Thus, practical reinforcement learning algorithms often incorporate a mixture of Bayesian and domain-specific ideas to bring down sample complexity and computational burden. Notable examples include guided policy search with unknown dynamics (Levine & Abbeel, 2014) and PILCO (Deisenroth & Rasmussen, 2011). These methods can learn a task using a few minutes to a few hours of real experience, compared to days or even weeks required by previous methods (Schulman et al., 2015; 2016; Lillicrap et al., 2015). However, these methods tend to make assumptions about the environment (e.g., instrumentation for access to the state at learning time), or become computationally intractable in high-dimensional settings (Wahlström et al., 2015).

Rather than hand-designing domain-specific reinforcement learning algorithms, we take a different approach in this paper: we view the learning process of the agent itself as an objective, which can be optimized using standard reinforcement learning algorithms. The objective is averaged across all possible MDPs according to a specific distribution, which reflects the prior that we would like to distill into the agent. We structure the agent as a recurrent neural network, which receives past rewards, actions, and termination flags as inputs in addition to the normally received observations. Furthermore, its internal state is preserved across episodes, so that it has the capacity to perform learning in its own hidden activations. The learned agent thus also acts as the learning algorithm, and can adapt to the task at hand when deployed.

We evaluate this approach on two sets of classical problems, multi-armed bandits and tabular MDPs. These problems have been extensively studied, and there exist algorithms that achieve asymptotically optimal performance. We demonstrate that our method, named RL$^2$, can achieve performance comparable with these theoretically justified algorithms. Next, we evaluate RL$^2$ on a vision-based navigation task implemented using the ViZDoom environment (Kempka et al., 2016), showing that RL$^2$ can also scale to high-dimensional problems.

## 2 METHOD

### 2.1 PRELIMINARIES

We define a discrete-time finite-horizon discounted Markov decision process (MDP) by a tuple $M = (\mathcal{S}, \mathcal{A}, \mathcal{P}, r, \rho_0, \gamma, T)$, in which $\mathcal{S}$ is a state set, $\mathcal{A}$ an action set, $\mathcal{P} : \mathcal{S} \times \mathcal{A} \times \mathcal{S} \to \mathbb{R}_+$ a transition probability distribution, $r : \mathcal{S} \times \mathcal{A} \to [-R_{\max}, R_{\max}]$ a bounded reward function, $\rho_0 : \mathcal{S} \to \mathbb{R}_+$ an initial state distribution, $\gamma \in [0, 1]$ a discount factor, and $T$ the horizon. In policy search methods, we typically optimize a stochastic policy $\pi_\theta : \mathcal{S} \times \mathcal{A} \to \mathbb{R}_+$ parametrized by $\theta$. The objective is to maximize its expected discounted return, $\eta(\pi_\theta) = \mathbb{E}_\tau[\sum_{t=0}^{T} \gamma^t r(s_t, a_t)]$, where $\tau = (s_0, a_0, \ldots)$ denotes the whole trajectory, $s_0 \sim \rho_0(s_0)$, $a_t \sim \pi_\theta(a_t|s_t)$, and $s_{t+1} \sim \mathcal{P}(s_{t+1}|s_t, a_t)$.

### 2.2 FORMULATION

We now describe our formulation, which casts learning an RL algorithm as a reinforcement learning problem, and hence the name RL$^2$. We assume knowledge of a set of MDPs, denoted by $\mathcal{M}$, and a distribution over them: $\rho_\mathcal{M} : \mathcal{M} \to \mathbb{R}_+$. We only need to sample from this distribution. We use $n$ to denote the total number of episodes allowed to spend with a specific MDP. We define a *trial* to be such a series of episodes of interaction with a fixed MDP.

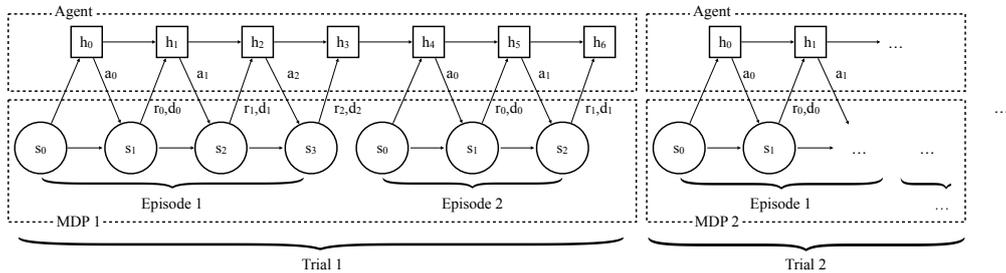

Figure 1: Procedure of agent-environment interaction

This process of interaction between an agent and the environment is illustrated in Figure 1. Here, each trial happens to consist of two episodes, hence $n = 2$. For each trial, a separate MDP is drawn from $\rho_\mathcal{M}$, and for each episode, a fresh $s_0$ is drawn from the initial state distribution specific to the corresponding MDP. Upon receiving an action $a_t$ produced by the agent, the environment computes reward $r_t$, steps forward, and computes the next state $s_{t+1}$. If the episode has terminated, it sets termination flag $d_t$ to 1, which otherwise defaults to 0. Together, the next state $s_{t+1}$, action $a_t$, reward $r_t$, and termination flag $d_t$, are concatenated to form the input to the policy[1], which, conditioned on the hidden state $h_{t+1}$, generates the next hidden state $h_{t+2}$ and action $a_{t+1}$. At the end of an episode, the hidden state of the policy is preserved to the next episode, but not preserved between trials.

The objective under this formulation is to maximize the expected total discounted reward accumulated during a single trial rather than a single episode. Maximizing this objective is equivalent to minimizing the cumulative pseudo-regret (Bubeck & Cesa-Bianchi, 2012). Since the underlying MDP changes across trials, as long as different strategies are required for different MDPs, the agent must act differently according to its belief over which MDP it is currently in. Hence, the agent is forced to integrate all the information it has received, including past actions, rewards, and termination flags, and adapt its strategy continually. Hence, we have set up an end-to-end optimization process, where the agent is encouraged to learn a "fast" reinforcement learning algorithm.

For clarity of exposition, we have defined the "inner" problem (of which the agent sees $n$ each trials) to be an MDP rather than a POMDP. However, the method can also be applied in the partially-observed setting without any conceptual changes. In the partially observed setting, the agent is faced with a sequence of POMDPs, and it receives an observation $o_t$ instead of state $s_t$ at time $t$. The visual navigation experiment in Section 3.3, is actually an instance of the this POMDP setting.

### 2.3 POLICY REPRESENTATION

We represent the policy as a general recurrent neural network. Each timestep, it receives the tuple $(s, a, r, d)$ as input, which is embedded using a function $\phi(s, a, r, d)$ and provided as input to an RNN. To alleviate the difficulty of training RNNs due to vanishing and exploding gradients (Bengio et al., 1994), we use Gated Recurrent Units (GRUs) (Cho et al., 2014) which have been demonstrated to have good empirical performance (Chung et al., 2014; Józefowicz et al., 2015). The output of the GRU is fed to a fully connected layer followed by a softmax function, which forms the distribution over actions.

We have also experimented with alternative architectures which explicitly reset part of the hidden state each episode of the sampled MDP, but we did not find any improvement over the simple architecture described above.

### 2.4 POLICY OPTIMIZATION

After formulating the task as a reinforcement learning problem, we can readily use standard off-the-shelf RL algorithms to optimize the policy. We use a first-order implementation of Trust Region Policy Optimization (TRPO) (Schulman et al., 2015), because of its excellent empirical performance, and because it does not require excessive hyperparameter tuning. For more details, we refer the reader to the original paper. To reduce variance in the stochastic gradient estimation, we use a baseline which is also represented as an RNN using GRUs as building blocks. We optionally apply Generalized Advantage Estimation (GAE) (Schulman et al., 2016) to further reduce the variance.

## 3 EVALUATION

We designed experiments to answer the following questions:

- Can RL[2] learn algorithms that achieve good performance on MDP classes with special structure, relative to existing algorithms tailored to this structure that have been proposed in the literature?
- Can RL[2] scale to high-dimensional tasks?

For the first question, we evaluate RL[2] on two sets of tasks, multi-armed bandits (MAB) and tabular MDPs. These problems have been studied extensively in the reinforcement learning literature, and this body of work includes algorithms with guarantees of asymptotic optimality. We demonstrate that our approach achieves comparable performance to these theoretically justified algorithms.

---

[1] To make sure that the inputs have a consistent dimension, we use placeholder values for the initial input to the policy.

For the second question, we evaluate RL² on a vision-based navigation task. Our experiments show that the learned policy makes effective use of the learned visual information and also short-term information acquired from previous episodes.

### 3.1 MULTI-ARMED BANDITS

Multi-armed bandit problems are a subset of MDPs where the agent's environment is stateless. Specifically, there are $k$ arms (actions), and at every time step, the agent pulls one of the arms, say $i$, and receives a reward drawn from an unknown distribution: our experiments take each arm to be a Bernoulli distribution with parameter $p_i$. The goal is to maximize the total reward obtained over a fixed number of time steps. The key challenge is balancing exploration and exploitation—"exploring" each arm enough times to estimate its distribution ($p_i$), but eventually switching over to "exploitation" of the best arm. Despite the simplicity of multi-arm bandit problems, their study has led to a rich theory and a collection of algorithms with optimality guarantees.

Using RL², we can train an RNN policy to solve bandit problems by training it on a given distribution $\rho_\mathcal{M}$. If the learning is successful, the resulting policy should be able to perform competitively with the theoretically optimal algorithms. We randomly generated bandit problems by sampling each parameter $p_i$ from the uniform distribution on $[0, 1]$. After training the RNN policy with RL², we compared it against the following strategies:

- Random: this is a baseline strategy, where the agent pulls a random arm each time.

- Gittins index (Gittins, 1979): this method gives the Bayes optimal solution in the discounted infinite-horizon case, by computing an index separately for each arm, and taking the arm with the largest index. While this work shows it is sufficient to independently compute an index for each arm (hence avoiding combinatorial explosion with the number of arms), it doesn't show how to tractably compute these individual indices exactly. We follow the practical approximations described in Gittins et al. (2011), Chakravorty & Mahajan (2013), and Whittle (1982), and choose the best-performing approximation for each setup.

- UCB1 (Auer, 2002): this method estimates an upper-confidence bound, and pulls the arm with the largest value of $\text{ucb}_i(t) = \hat{\mu}_i(t-1) + c\sqrt{\frac{2\log t}{T_i(t-1)}}$, where $\hat{\mu}_i(t-1)$ is the estimated mean parameter for the $i$th arm, $T_i(t-1)$ is the number of times the $i$th arm has been pulled, and $c$ is a tunable hyperparameter (Audibert & Munos, 2011). We initialize the statistics with exactly one success and one failure, which corresponds to a $\text{Beta}(1, 1)$ prior.

- Thompson sampling (TS) (Thompson, 1933): this is a simple method which, at each time step, samples a list of arm means from the posterior distribution, and choose the best arm according to this sample. It has been demonstrated to compare favorably to UCB1 empirically (Chapelle & Li, 2011). We also experiment with an optimistic variant (OTS) (May et al., 2012), which samples $N$ times from the posterior, and takes the one with the highest probability.

- $\epsilon$-Greedy: in this strategy, the agent chooses the arm with the best empirical mean with probability $1 - \epsilon$, and chooses a random arm with probability $\epsilon$. We use the same initialization as UCB1.

- Greedy: this is a special case of $\epsilon$-Greedy with $\epsilon = 0$.

The Bayesian methods, Gittins index and Thompson sampling, take advantage of the distribution $\rho_\mathcal{M}$; and we provide these methods with the true distribution. For each method with hyperparameters, we maximize the score with a separate grid search for each of the experimental settings. The hyperparameters used for TRPO are shown in the appendix.

The results are summarized in Table 1. Learning curves for various settings are shown in Figure 2. We observe that our approach achieves performance that is almost as good as the the reference methods, which were (human) designed specifically to perform well on multi-armed bandit problems. It is worth noting that the published algorithms are mostly designed to minimize asymptotic regret (rather than finite horizon regret), hence there tends to be a little bit of room to outperform them in the finite horizon settings.

Table 1: MAB Results. Each grid cell records the total reward averaged over 1000 different instances of the bandit problem. We consider $k \in \{5, 10, 50\}$ bandits and $n \in \{10, 100, 500\}$ episodes of interaction. We highlight the best-performing algorithms in each setup according to the computed mean, and we also highlight the other algorithms in that row whose performance is not significantly different from the best one (determined by a one-sided $t$-test with $p = 0.05$).

| Setup | Random | Gittins | TS | OTS | UCB1 | $\epsilon$-Greedy | Greedy | RL$^2$ |
|---|---|---|---|---|---|---|---|---|
| $n = 10, k = 5$ | 5.0 | **6.6** | 5.7 | 6.5 | **6.7** | **6.6** | **6.6** | **6.7** |
| $n = 10, k = 10$ | 5.0 | **6.6** | 5.5 | 6.2 | **6.7** | **6.6** | **6.6** | **6.7** |
| $n = 10, k = 50$ | 5.1 | 6.5 | 5.2 | 5.5 | **6.6** | 6.5 | 6.5 | **6.8** |
| $n = 100, k = 5$ | 49.9 | **78.3** | 74.7 | **77.9** | **78.0** | 75.4 | 74.8 | **78.7** |
| $n = 100, k = 10$ | 49.9 | **82.8** | 76.7 | 81.4 | **82.4** | 77.4 | 77.1 | **83.5** |
| $n = 100, k = 50$ | 49.8 | **85.2** | 64.5 | 67.7 | **84.3** | 78.3 | 78.0 | **84.9** |
| $n = 500, k = 5$ | 249.8 | **405.8** | 402.0 | **406.7** | **405.8** | 388.2 | 380.6 | **401.6** |
| $n = 500, k = 10$ | 249.0 | **437.8** | 429.5 | **438.9** | **437.1** | 408.0 | 395.0 | 432.5 |
| $n = 500, k = 50$ | 249.6 | **463.7** | 427.2 | 437.6 | 457.6 | 413.6 | 402.8 | 438.9 |

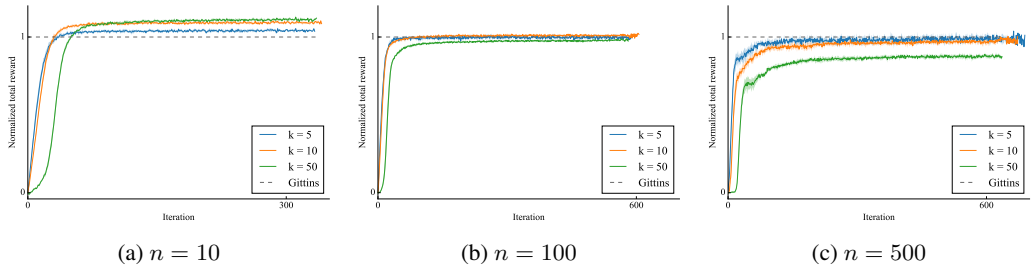

(a) $n = 10$  (b) $n = 100$  (c) $n = 500$

Figure 2: RL$^2$ learning curves for multi-armed bandits. Performance is normalized such that Gittins index scores 1, and random policy scores 0.

We observe that there is a noticeable gap between Gittins index and RL$^2$ in the most challenging scenario, with 50 arms and 500 episodes. This raises the question whether better architectures or better (slow) RL algorithms should be explored. To determine the bottleneck, we trained the same policy architecture using supervised learning, using the trajectories generated by the Gittins index approach as training data. We found that the learned policy, when executed in test domains, achieved the same level of performance as the Gittins index approach, suggesting that there is room for improvement by using better RL algorithms.

### 3.2 TABULAR MDPS

The bandit problem provides a natural and simple setting to investigate whether the policy learns to trade off between exploration and exploitation. However, the problem itself involves no sequential decision making, and does not fully characterize the challenges in solving MDPs. Hence, we perform further experiments using randomly generated tabular MDPs, where there is a finite number of possible states and actions—small enough that the transition probability distribution can be explicitly given as a table. We compare our approach with the following methods:

- Random: the agent chooses an action uniformly at random for each time step;
- PSRL (Strens, 2000; Osband et al., 2013): this is a direct generalization of Thompson sampling to MDPs, where at the beginning of each episode, we sample an MDP from the posterior distribution, and take actions according to the optimal policy for the entire episode. Similarly, we include an optimistic variant (OPSRL), which has also been explored in Osband & Van Roy (2016).
- BEB (Kolter & Ng, 2009): this is a model-based optimistic algorithm that adds an exploration bonus to (thus far) infrequently visited states and actions.

- UCRL2 (Jaksch et al., 2010): this algorithm computes, at each iteration, the optimal policy against an optimistic MDP under the current belief, using an extended value iteration procedure.
- $\epsilon$-Greedy: this algorithm takes actions optimal against the MAP estimate according to the current posterior, which is updated once per episode.
- Greedy: a special case of $\epsilon$-Greedy with $\epsilon = 0$.

Table 2: Random MDP Results

| **Setup** | **Random** | **PSRL** | **OPSRL** | **UCRL2** | **BEB** | **$\epsilon$-Greedy** | **Greedy** | **RL$^2$** |
|---|---|---|---|---|---|---|---|---|
| $n = 10$ | 100.1 | 138.1 | 144.1 | 146.6 | 150.2 | 132.8 | 134.8 | **156.2** |
| $n = 25$ | 250.2 | 408.8 | 425.2 | 424.1 | 427.8 | 377.3 | 368.8 | **445.7** |
| $n = 50$ | 499.7 | 904.4 | **930.7** | 918.9 | 917.8 | 823.3 | 769.3 | **936.1** |
| $n = 75$ | 749.9 | 1417.1 | **1449.2** | 1427.6 | 1422.6 | 1293.9 | 1172.9 | 1428.8 |
| $n = 100$ | 999.4 | 1939.5 | **1973.9** | 1942.1 | 1935.1 | 1778.2 | 1578.5 | 1913.7 |

The distribution over MDPs is constructed with $|\mathcal{S}| = 10$, $|\mathcal{A}| = 5$. The rewards follow a Gaussian distribution with unit variance, and the mean parameters are sampled independently from $\text{Normal}(1, 1)$. The transitions are sampled from a flat Dirichlet distribution. This construction matches the commonly used prior in Bayesian RL methods. We set the horizon for each episode to be $T = 10$, and an episode always starts on the first state.

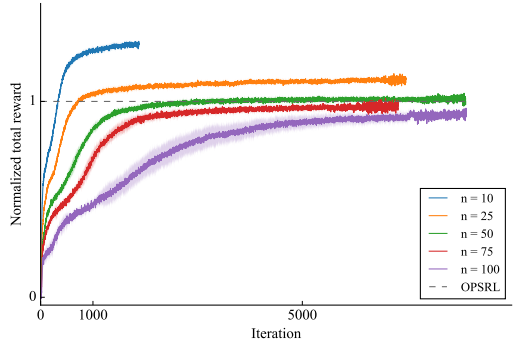

Figure 3: RL$^2$ learning curves for tabular MDPs. Performance is normalized such that OPSRL scores 1, and random policy scores 0.

The results are summarized in Table 2, and the learning curves are shown in Figure 3. We follow the same evaluation procedure as in the bandit case. We experiment with $n \in \{10, 25, 50, 75, 100\}$. For fewer episodes, our approach surprisingly outperforms existing methods by a large margin. The advantage is reversed as $n$ increases, suggesting that the reinforcement learning problem in the outer loop becomes more challenging to solve. We think that the advantage for small $n$ comes from the need for more aggressive exploitation: since there are 140 degrees of freedom to estimate in order to characterize the MDP, and by the 10th episode, we will not have enough samples to form a good estimate of the entire dynamics. By directly optimizing the RNN in this setting, our approach should be able to cope with this shortage of samples, and decides to exploit sooner compared to the reference algorithms.

### 3.3 VISUAL NAVIGATION

The previous two tasks both only involve very low-dimensional state spaces. To evaluate the feasibility of scaling up RL$^2$, we further experiment with a challenging vision-based task, where the agent is asked to navigate a randomly generated maze to find a randomly placed target[2]. The agent receives a $+1$ reward when it reaches the target, $-0.001$ when it hits the wall, and $-0.04$ per time step to encourage it to reach targets faster. It can interact with the maze for multiple episodes, during which the maze structure and target position are held fixed. The optimal strategy is to explore the maze efficiently during the first episode, and after locating the target, act optimally against the current maze and target based on the collected information. An illustration of the task is given in Figure 4.

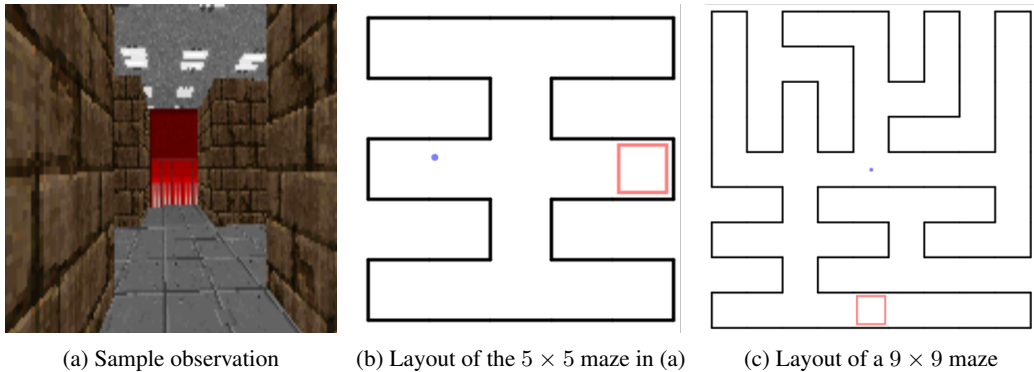

(a) Sample observation     (b) Layout of the $5 \times 5$ maze in (a)     (c) Layout of a $9 \times 9$ maze

Figure 4: Visual navigation. The target block is shown in red, and occupies an entire grid in the maze layout.

Visual navigation alone is a challenging task for reinforcement learning. The agent only receives very sparse rewards during training, and does not have the primitives for efficient exploration at the beginning of training. It also needs to make efficient use of memory to decide how it should explore the space, without forgetting about where it has already explored. Previously, Oh et al. (2016) have studied similar vision-based navigation tasks in Minecraft. However, they use higher-level actions for efficient navigation. Similar high-level actions in our task would each require around 5 low-level actions combined in the right way. In contrast, our RL$^2$ agent needs to learn these higher-level actions from scratch.

We use a simple training setup, where we use small mazes of size $5 \times 5$, with 2 episodes of interaction, each with horizon up to 250. Here the size of the maze is measured by the number of grid cells along each wall in a discrete representation of the maze. During each trial, we sample 1 out of 1000 randomly generated configurations of map layout and target positions. During testing, we evaluate on 1000 separately generated configurations. In addition, we also study its extrapolation behavior along two axes, by (1) testing on large mazes of size $9 \times 9$ (see Figure 4c) and (2) running the agent for up to 5 episodes in both small and large mazes. For the large maze, we also increase the horizon per episode by 4x due to the increased size of the maze.

Table 3: Results for visual navigation. These metrics are computed using the best run among all runs shown in Figure 5. In 3c, we measure the proportion of mazes where the trajectory length in the second episode does not exceed the trajectory length in the first episode.

| (a) Average length of successful trajectories | | | (b) %Success | | | (c) %Improved | |
|---|---|---|---|---|---|---|---|
| **Episode** | **Small** | **Large** | **Episode** | **Small** | **Large** | **Small** | **Large** |
| 1 | $52.4 \pm 1.3$ | $180.1 \pm 6.0$ | 1 | 99.3% | 97.1% | 91.7% | 71.4% |
| 2 | $39.1 \pm 0.9$ | $151.8 \pm 5.9$ | 2 | 99.6% | 96.7% | | |
| 3 | $42.6 \pm 1.0$ | $169.3 \pm 6.3$ | 3 | 99.7% | 95.8% | | |
| 4 | $43.5 \pm 1.1$ | $162.3 \pm 6.4$ | 4 | 99.4% | 95.6% | | |
| 5 | $43.9 \pm 1.1$ | $169.3 \pm 6.5$ | 5 | 99.6% | 96.1% | | |

---

[2]Videos for the task are available at https://goo.gl/rDDBpb.

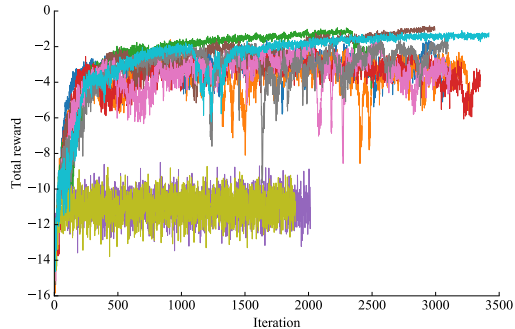

Figure 5: RL$^2$ learning curves for visual navigation. Each curve shows a different random initialization of the RNN weights. Performance varies greatly across different initializations.

The results are summarized in Table 3, and the learning curves are shown in Figure 5. We observe that there is a significant reduction in trajectory lengths between the first two episodes in both the smaller and larger mazes, suggesting that the agent has learned how to use information from past episodes. It also achieves reasonable extrapolation behavior in further episodes by maintaining its performance, although there is a small drop in the rate of success in the larger mazes. We also observe that on larger mazes, the ratio of improved trajectories is lower, likely because the agent has not learned how to act optimally in the larger mazes.

Still, even on the small mazes, the agent does not learn to perfectly reuse prior information. An illustration of the agent's behavior is shown in Figure 6. The intended behavior, which occurs most frequently, as shown in 6a and 6b, is that the agent should remember the target's location, and utilize it to act optimally in the second episode. However, occasionally the agent forgets about where the target was, and continues to explore in the second episode, as shown in 6c and 6d. We believe that better reinforcement learning techniques used as the outer-loop algorithm will improve these results in the future.

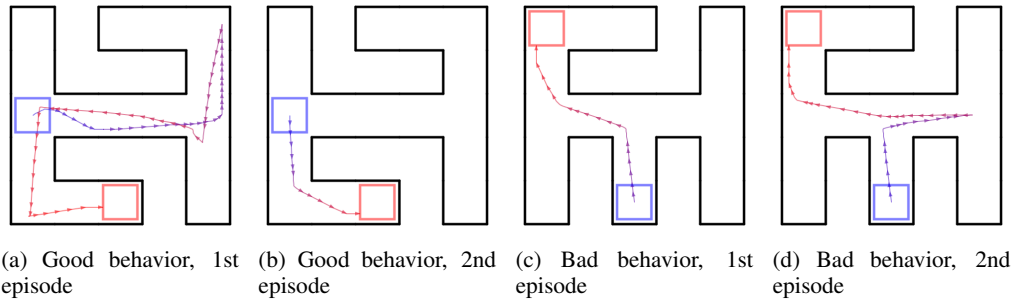

(a) Good behavior, 1st episode  (b) Good behavior, 2nd episode  (c) Bad behavior, 1st episode  (d) Bad behavior, 2nd episode

Figure 6: Visualization of the agent's behavior. In each scenario, the agent starts at the center of the blue block, and the goal is to reach anywhere in the red block.

## 4 RELATED WORK

The concept of using prior experience to speed up reinforcement learning algorithms has been explored in the past in various forms. Earlier studies have investigated automatic tuning of hyperparameters, such as learning rate and temperature (Ishii et al., 2002; Schweighofer & Doya, 2003), as a form of meta-learning. Wilson et al. (2007) use hierarchical Bayesian methods to maintain a posterior over possible models of dynamics, and apply optimistic Thompson sampling according to the posterior. Many works in hierarchical reinforcement learning propose to extract reusable skills from previous tasks to speed up exploration in new tasks (Singh, 1992; Perkins et al., 1999). We refer the reader to Taylor & Stone (2009) for a more thorough survey on the multi-task and transfer learning aspects.

More recently, Fu et al. (2015) propose a model-based approach on top of iLQG with unknown dynamics (Levine & Abbeel, 2014), which uses samples collected from previous tasks to build a neural network prior for the dynamics, and can perform one-shot learning on new, but related tasks thanks to reduced sample complexity. There has been a growing interest in using deep neural networks for multi-task learning and transfer learning (Parisotto et al., 2015; Rusu et al., 2015; 2016a; Devin et al., 2016; Rusu et al., 2016b).

In the broader context of machine learning, there has been a lot of interest in one-shot learning for object classification (Vilalta & Drissi, 2002; Fei-Fei et al., 2006; Larochelle et al., 2008; Lake et al., 2011; Koch, 2015). Our work draws inspiration from a particular line of work (Younger et al., 2001; Santoro et al., 2016; Vinyals et al., 2016), which formulates meta-learning as an optimization problem, and can thus be optimized end-to-end via gradient descent. While these work applies to the supervised learning setting, our work applies in the more general reinforcement learning setting. Although the reinforcement learning setting is more challenging, the resulting behavior is far richer: our agent must not only learn to exploit existing information, but also learn to explore, a problem that is usually not a factor in supervised learning. Another line of work (Hochreiter et al., 2001; Younger et al., 2001; Andrychowicz et al., 2016; Li & Malik, 2016) studies meta-learning over the optimization process. There, the meta-learner makes explicit updates to a parametrized model. In comparison, we do not use a directly parametrized policy; instead, the recurrent neural network agent acts as the meta-learner and the resulting policy simultaneously.

Our formulation essentially constructs a partially observable MDP (POMDP) which is solved in the outer loop, where the underlying MDP is unobserved by the agent. This reduction of an unknown MDP to a POMDP can be traced back to dual control theory (Feldbaum, 1960), where "dual" refers to the fact that one is controlling both the state and the state estimate. Feldbaum pointed out that the solution can in principle be computed with dynamic programming, but doing so is usually impractical. POMDPs with such structure have also been studied under the name "mixed observability MDPs" (Ong et al., 2010). However, the method proposed there suffers from the usual challenges of solving POMDPs in high dimensions.

## 5 DISCUSSION

This paper suggests a different approach for designing better reinforcement learning algorithms: instead of acting as the designers ourselves, learn the algorithm end-to-end using standard reinforcement learning techniques. That is, the "fast" RL algorithm is a computation whose state is stored in the RNN activations, and the RNN's weights are learned by a general-purpose "slow" reinforcement learning algorithm. Our method, $RL^2$, has demonstrated competence comparable with theoretically optimal algorithms in small-scale settings. We have further shown its potential to scale to high-dimensional tasks.

In the experiments, we have identified opportunities to improve upon $RL^2$: the outer-loop reinforcement learning algorithm was shown to be an immediate bottleneck, and we believe that for settings with extremely long horizons, better architecture may also be required for the policy. Although we have used generic methods and architectures for the outer-loop algorithm and the policy, doing this also ignores the underlying episodic structure. We expect algorithms and policy architectures that exploit the problem structure to significantly boost the performance.

### ACKNOWLEDGMENTS

We would like to thank our colleagues at Berkeley and OpenAI for insightful discussions. This research was funded in part by ONR through a PECASE award. Yan Duan was also supported by a Berkeley AI Research lab Fellowship and a Huawei Fellowship. Xi Chen was also supported by a Berkeley AI Research lab Fellowship. We gratefully acknowledge the support of the NSF through grant IIS-1619362 and of the ARC through a Laureate Fellowship (FL110100281) and through the ARC Centre of Excellence for Mathematical and Statistical Frontiers.

Shin Ishii, Wako Yoshida, and Junichiro Yoshimoto. Control of exploitation–exploration meta-parameter in reinforcement learning. *Neural networks*, 15(4):665–687, 2002.

Thomas Jaksch, Ronald Ortner, and Peter Auer. Near-optimal regret bounds for reinforcement learning. *Journal of Machine Learning Research*, 11(Apr):1563–1600, 2010.

Rafal Józefowicz, Wojciech Zaremba, and Ilya Sutskever. An empirical exploration of recurrent network architectures. In *Proceedings of the 32nd International Conference on Machine Learning, ICML 2015, Lille, France, 6-11 July 2015*, pp. 2342–2350, 2015. URL http://jmlr.org/proceedings/papers/v37/jozefowicz15.html.

Michał Kempka, Marek Wydmuch, Grzegorz Runc, Jakub Toczek, and Wojciech Jaśkowski. Vizdoom: A doom-based ai research platform for visual reinforcement learning. *arXiv preprint arXiv:1605.02097*, 2016.

Gregory Koch. *Siamese neural networks for one-shot image recognition*. PhD thesis, University of Toronto, 2015.

J Zico Kolter and Andrew Y Ng. Near-bayesian exploration in polynomial time. In *Proceedings of the 26th Annual International Conference on Machine Learning*, pp. 513–520. ACM, 2009.

Brenden M Lake, Ruslan Salakhutdinov, Jason Gross, and Joshua B Tenenbaum. One shot learning of simple visual concepts. In *Proceedings of the 33rd Annual Conference of the Cognitive Science Society*, volume 172, pp. 2, 2011.

Hugo Larochelle, Dumitru Erhan, and Yoshua Bengio. Zero-data learning of new tasks. In *AAAI*, volume 1, pp. 3, 2008.

Sergey Levine and Pieter Abbeel. Learning neural network policies with guided policy search under unknown dynamics. In *Advances in Neural Information Processing Systems*, pp. 1071–1079, 2014.

Sergey Levine, Chelsea Finn, Trevor Darrell, and Pieter Abbeel. End-to-end training of deep visuomotor policies. *Journal of Machine Learning Research*, 17(39):1–40, 2016.

Ke Li and Jitendra Malik. Learning to optimize. *arXiv preprint arXiv:1606.01885*, 2016.

Timothy P Lillicrap, Jonathan J Hunt, Alexander Pritzel, Nicolas Heess, Tom Erez, Yuval Tassa, David Silver, and Daan Wierstra. Continuous control with deep reinforcement learning. *arXiv preprint arXiv:1509.02971*, 2015.

Benedict C May, Nathan Korda, Anthony Lee, and David S Leslie. Optimistic bayesian sampling in contextual-bandit problems. *Journal of Machine Learning Research*, 13(Jun):2069–2106, 2012.

Volodymyr Mnih, Koray Kavukcuoglu, David Silver, Andrei A Rusu, Joel Veness, Marc G Bellemare, Alex Graves, Martin Riedmiller, Andreas K Fidjeland, Georg Ostrovski, et al. Human-level control through deep reinforcement learning. *Nature*, 518(7540):529–533, 2015.

Junhyuk Oh, Valliappa Chockalingam, Satinder Singh, and Honglak Lee. Control of memory, active perception, and action in minecraft. *arXiv preprint arXiv:1605.09128*, 2016.

Sylvie CW Ong, Shao Wei Png, David Hsu, and Wee Sun Lee. Planning under uncertainty for robotic tasks with mixed observability. *The International Journal of Robotics Research*, 29(8): 1053–1068, 2010.

Ian Osband and Benjamin Van Roy. Why is posterior sampling better than optimism for reinforcement learning. *arXiv preprint arXiv:1607.00215*, 2016.

Ian Osband, Dan Russo, and Benjamin Van Roy. (more) efficient reinforcement learning via posterior sampling. In *Advances in Neural Information Processing Systems*, pp. 3003–3011, 2013.

Emilio Parisotto, Jimmy Lei Ba, and Ruslan Salakhutdinov. Actor-mimic: Deep multitask and transfer reinforcement learning. *arXiv preprint arXiv:1511.06342*, 2015.11

APPENDIX

A   DETAILED EXPERIMENT SETUP

Common to all experiments: as mentioned in Section 2.2, we use placeholder values when necessary. For example, at $t = 0$ there is no previous action, reward, or termination flag. Since all of our experiments use discrete actions, we use the embedding of the action 0 as a placeholder for actions, and 0 for both the rewards and termination flags. To form the input to the GRU, we use the values for the rewards and termination flags as-is, and embed the states and actions as described separately below for each experiments. These values are then concatenated together to form the joint embedding.

For the neural network architecture, We use rectified linear units throughout the experiments as the hidden activation, and we apply weight normalization without data-dependent initialization (Salimans & Kingma, 2016) to all weight matrices. The hidden-to-hidden weight matrix uses an orthogonal initialization (Saxe et al., 2013), and all other weight matrices use Xavier initialization (Glorot & Bengio, 2010). We initialize all bias vectors to 0. Unless otherwise mentioned, the policy and the baseline uses separate neural networks with the same architecture until the final layer, where the number of outputs differ.

All experiments are implemented using TensorFlow (Abadi et al., 2016) and rllab (Duan et al., 2016). We use the implementations of classic algorithms provided by the TabulaRL package (Osband, 2016).

A.1   MULTI-ARMED BANDITS

The parameters for TRPO are shown in Table 1. Since the environment is stateless, we use a constant embedding 0 as a placeholder in place of the states, and a one-hot embedding for the actions.

Table 1: Hyperparameters for TRPO: multi-armed bandits

| Discount | 0.99 |
| --- | --- |
| GAE $\lambda$ | 0.3 |
| Policy Iters | Up to 1000 |
| #GRU Units | 256 |
| Mean KL | 0.01 |
| Batch size | 250000 |

A.2   TABULAR MDPS

The parameters for TRPO are shown in Table 2. We use a one-hot embedding for the states and actions separately, which are then concatenated together.

Table 2: Hyperparameters for TRPO: tabular MDPs

| Discount | 0.99 |
| --- | --- |
| GAE $\lambda$ | 0.3 |
| Policy Iters | Up to 10000 |
| #GRU Units | 256 |
| Mean KL | 0.01 |
| Batch size | 250000 |

A.3   VISUAL NAVIGATION

The parameters for TRPO are shown in Table 3. For this task, we use a neural network to form the joint embedding. We rescale the images to have width 40 and height 30 with RGB channels preserved, and we recenter the RGB values to lie within range $[-1, 1]$. Then, this preprocessed image is passed through 2 convolution layers, each with 16 filters of size $5 \times 5$ and stride 2. The action is first embedded into a 256-dimensional vector where the embedding is learned, and then concatenated with the flattened output of the final convolution layer. The joint vector is then fed to a fully connected layer with 256 hidden units.

Unlike previous experiments, we let the policy and the baseline share the same neural network. We found this to improve the stability of training baselines and also the end performance of the policy, possibly due to regularization effects and better learned features imposed by weight sharing. Similar weight-sharing techniques have also been explored in (Mnih et al., 2016).

Table 3: Hyperparameters for TRPO: visual navigation

| | |
|---|---|
| Discount | 0.99 |
| GAE $\lambda$ | 0.99 |
| Policy Iters | Up to 5000 |
| #GRU Units | 256 |
| Mean KL | 0.01 |
| Batch size | 50000 |